\documentclass[10pt,journal,compsoc]{IEEEtran}
\usepackage{amsmath,amsfonts}
\usepackage{algorithmic}
\usepackage{algorithm}
\usepackage{array}
\usepackage{textcomp}
\usepackage{stfloats}
\usepackage{url}
\usepackage{verbatim}
\usepackage{graphicx}

\usepackage{amssymb, graphicx, amsmath, epstopdf}
\usepackage{amsthm}
\usepackage{cases}
\usepackage{amsfonts,amsmath,amsthm,amsfonts,amssymb}
\usepackage{amssymb,amsfonts,amsthm}
\usepackage{graphicx}
\usepackage{hyperref}
\usepackage{cleveref}
\usepackage{aliascnt}
\usepackage[utf8]{inputenc}
\usepackage{xcolor}         

\usepackage{bbm}
\usepackage{algorithm}
\usepackage{algorithmic}
\usepackage{caption}
\usepackage{subcaption}
\usepackage{bm}
\usepackage{empheq}
\usepackage{multirow}
\usepackage{ragged2e}
\allowdisplaybreaks[4]
\usepackage[backend=bibtex, style=ieee, giveninits=true, doi=false]{biblatex}
\newaliascnt{remark}{theorem}
\newtheorem{remark}[remark]{Remark}
\aliascntresetthe{remark}
\crefname{remark}{remark}{remarks}
\Crefname{remark}{Remark}{Remarks}

\begin{document}

\title{Approximate and Weighted Data Reconstruction Attack in Federated Learning}

\author{Yongcun Song, Ziqi Wang, and Enrique Zuazua
	\IEEEcompsocitemizethanks{
		\IEEEcompsocthanksitem Y. Song is with Division of Mathematical Sciences, School of Physical and Mathematical Sciences, Nanyang Technological University, 21 Nanyang Link, 637371, Singapore. Email: yongcun.song@ntu.edu.sg.
		\IEEEcompsocthanksitem Z. Wang (corresponding author) is with the Chair for Dynamics, Control, Machine Learning and Numerics – Alexander von Humboldt Professorship, Department of Mathematics, Friedrich-Alexander-Universit\"at Erlangen-N\"urnberg, Cauerstrasse 11, 91058 Erlangen, Germany. Email: ziqi.wang@fau.de.
		\IEEEcompsocthanksitem E. Zuazua is with the Chair for Dynamics, Control, Machine Learning and Numerics – Alexander von Humboldt Professorship, Department of Mathematics, Friedrich-Alexander-Universit\"at Erlangen-N\"urnberg, Cauerstrasse 11, 91058 Erlangen, Germany; the Chair of Computational Mathematics, Fundaci\'on Deusto Avda. de las Universidades 24, 48007 Bilbao, Basque Country, Spain; and also with the Departamento de Matem\'aticas, Universidad Aut\'onoma de Madrid, 28049 Madrid, Spain. Email: enrique.zuazua@fau.de.
		\IEEEcompsocthanksitem Authors are listed alphabetically by family name.
		\IEEEcompsocthanksitem Published in \textit{IEEE Transactions on Big Data}, vol. 12, no. 5, pp. 1606--1617, 2026. DOI: \href{https://doi.org/10.1109/TBDATA.2026.3695416}{10.1109/TBDATA.2026.3695416}.

		}
}

\markboth{}%
{Song \MakeLowercase{\textit{et al.}}: Approximate and Weighted Attack in FL}


\IEEEtitleabstractindextext{
	\begin{abstract}
		\justifying
		Federated learning (FL) is a distributed learning paradigm that enables multiple clients to collaboratively train a machine learning model without sharing private data. Although FL is often regarded as privacy-preserving by design, recent data reconstruction attacks demonstrate that adversaries can recover clients' training data from shared model updates. However, existing attack methods frequently fail in the widely used federated averaging (FedAvg) setting, where clients transmit model parameters only after executing multiple local training steps. To overcome this limitation, we propose an interpolation-based approximation method that makes attacks on FedAvg feasible by effectively estimating the intermediate model updates generated during local training. Furthermore, we design a layer-wise weighted loss function to enhance reconstruction quality. Specifically, weights are assigned to different layers based on the neural network architecture and are systematically tuned via Bayesian optimization. Experimental results demonstrate that the proposed approximate and weighted attack method outperforms existing state-of-the-art approaches, yielding substantial improvements across multiple image reconstruction metrics.
	\end{abstract}

	}

\maketitle

\IEEEdisplaynontitleabstractindextext

\IEEEpeerreviewmaketitle

\section{Introduction}

\IEEEPARstart{W}{ith} the escalating volume of data generated by distributed personal electronic devices, conventional centralized approaches for training machine learning models face substantial obstacles in data collection and privacy preservation. To mitigate these issues, federated learning (FL) \cite{li2020federated,mcmahan2017communicationefficient} has emerged as a promising paradigm.

A prominent feature of FL is that it enables model training on distributed data sources while retaining raw data locally. For instance, in the widely adopted federated averaging (FedAvg) \cite{mcmahan2017communicationefficient} algorithm, each client trains a local model on private data and transmits the updated model parameters to the server. The central server subsequently aggregates these updates to refine a global model. In this manner, FL allows multiple participants to train a robust model without directly exposing raw data, thereby addressing concerns related to data privacy, data security, and data access rights. Consequently, FL has gained significant attention in recent years for addressing growing privacy concerns in applications such as healthcare \cite{brisimi2018federated, xu2021federated} and cooperative controller learning across autonomous vehicles without sharing historical trajectories \cite{zeng2022federated}.

Although model updates in FL were long believed to be safe to share, recent studies \cite{geng2023improved,xu2022agic,yin2021see,zhu2019deep} have shown that clients' sensitive training data can be compromised through \emph{data reconstruction attacks} \cite{zhu2019deep}. In these attacks, the adversary first randomly initializes dummy samples and labels and then performs forward and backward propagation to obtain dummy model updates. Through an iterative process that minimizes the discrepancy between the dummy model updates and the ground-truth updates, the dummy samples and labels are updated simultaneously.

Existing attack methods have achieved promising results, but attacking FedAvg with multi-step model updates remains challenging. A more detailed discussion of the related literature is provided in \Cref{sec_related_work}. In particular, existing attacks are significantly less effective in the realistic FedAvg regime with multiple local epochs and multiple mini-batches, where the server observes only the final accumulated model update rather than the hidden epoch-wise updates. This missing-dynamics issue makes the corresponding reconstruction problem largely intractable for existing methods.

To address these challenges, we propose a novel approximate and weighted attack (AWA) method for data reconstruction against FL systems with multi-step model updates. The key contribution of this work is to transform an intractable reconstruction problem in realistic FedAvg settings (multi-epoch and multi-batch) into a tractable one through an interpolation-based approximation of hidden training dynamics. Concretely, we approximate the missing intermediate model updates by interpolating between the leaked final model parameters and the initial model parameters. Furthermore, we enhance the reconstruction process by using a layer-wise weighted loss whose weights are tuned systematically by Bayesian optimization \cite{frazier2018tutorial}.

Overall, our main contributions are as follows:
\begin{enumerate}
	\item To attack FL systems with multi-step model updates, we propose an interpolation-based approximation method. The model update associated with each epoch is approximated by interpolating between the initial and updated model parameters, which makes attacks on realistic FedAvg scenarios feasible and effective.
	\item To further improve attack performance after approximation, we employ a layer-wise weighted loss function. The layer weights are determined by Bayesian optimization, and we further enhance the weights of layers with large matching errors to improve the attack's adaptability and performance.
	\item Our method is architecture-aware through its layer-wise weighting design rather than being tied to a specific network structure. Furthermore, it can reconstruct training data from model updates leaked at different stages of the training process.
\end{enumerate}

The rest of the paper is organized as follows. In \Cref{sec_related_work}, we review the related literature on data reconstruction attacks in FL. In \Cref{sec_preliminaries}, we provide the necessary background on FL and data reconstruction attacks. In \Cref{sec_different_EB}, we analyze data reconstruction under different FedAvg settings. \Cref{sec_method} presents our proposed AWA method, including the approximation method and the layer-wise weighted loss function. The experimental setup and simulation results are presented in \Cref{sec_simulation}. Finally, \Cref{sec_conclusions} concludes the paper.

\section{Related Work}
\label{sec_related_work}

\noindent\textbf{Label Inference Techniques.} Inferring labels in advance avoids the joint optimization of both samples and labels, thereby markedly simplifying the reconstruction problem. It was first shown in \cite{zhao2020idlg} that the label of a single sample can be analytically extracted from the model updates of the last fully connected layer. Later, \cite{yin2021see} extended single-sample label extraction to batch settings under the restrictive assumption that labels do not repeat within a batch. This limitation was systematically addressed by the batch label inference approach proposed in \cite{geng2023improved}.

\noindent\textbf{Distance Functions for Matching Updates.} To measure the discrepancy between dummy and ground-truth model updates, Euclidean distance is commonly used in the attack loss function; see \cite{yin2021see,zhao2020idlg,zhu2019deep}. Moreover, cosine distance was advocated in \cite{geiping2020inverting,xu2022agic} because, in high dimensions, the direction of the model updates can carry more information than their magnitude. In \cite{wang2020sapag}, a Gaussian kernel based on model-update differences was proposed to measure the discrepancy, allowing the scaling factor in the kernel to adapt to the distribution of model updates in each attack.

\noindent\textbf{Optimizer Choices.} Among the optimizers used in these attacks, L-BFGS \cite{liu1989limited} and Adam \cite{kingma2014adam} are the most common. In particular, the reconstruction performance of these two optimizers was compared in \cite{geiping2020inverting, wang2020sapag,wei2020framework}. It was shown in \cite{wei2020framework} that L-BFGS requires fewer attack iterations than Adam to achieve high reconstruction quality on the LFW dataset \cite{huang2008labeled}. Conversely, \cite{geiping2020inverting} reported that L-BFGS obtains lower reconstruction quality than Adam on the CIFAR-10 dataset \cite{krizhevsky2009learning}. Although there is no universal rule for choosing the loss function or optimizer, appropriate selections can substantially improve attack performance in specific scenarios.

\noindent\textbf{Regularization Based on Prior Knowledge.} Another way to improve reconstruction performance is to add auxiliary regularization terms to the attack loss function based on prior knowledge of the data. In \cite{wei2020framework}, a label regularizer was proposed to match the dummy samples and labels when both are optimized simultaneously.
Moreover, image priors can be incorporated into the reconstruction procedure. For instance, total variation regularization \cite{geiping2020inverting} is used to reduce image noise. In \cite{yin2021see}, the authors proposed a prior term based on the statistical mean and variance stored in the batch normalization layers.
In \cite{jeon2021gradient}, a generative model pre-trained on the raw data distribution was used to improve reconstruction. However, the raw data distribution is not necessarily accessible during training.

\noindent\textbf{Attacks on Multi-Step FedAvg Updates.} Despite remarkable progress, limited attention has been paid to attacking FedAvg with multi-step updates, where clients share local model parameters only after training for multiple epochs, each executed over multiple mini-batches. GENG \cite{geng2023improved} extended gradient inversion attacks to FedAvg by modeling the unknown updates along an estimated trajectory between the initial and uploaded client models. However, its setting focuses on full-batch local training and does not consider layer-wise weighting. Another approximation method, AGIC \cite{xu2022agic}, first initializes a combined dummy batch whose size is the sum of all mini-batches used in the client's local training process. This combined dummy batch is then used to perform a single gradient step, and the resulting dummy model update is used to approximate the ground-truth multi-step model update. AGIC employs a weighted loss function, but the layer weights are chosen empirically rather than systematically.

\section{Preliminaries}
\label{sec_preliminaries}

In this section, we first describe the mathematical formulation and training process of FL and then introduce the formulation and setup of data reconstruction attacks.

\subsection{Problem Statement of FL}
In FL, the training data are distributed across $K$ clients, and the global training objective can be formulated as the following minimization problem \cite{li2020federated}:
\begin{equation}\label{eq_FL}
	\min_{\theta \in \mathbb{R}^{d}} \sum_{k=1}^K \rho_k F_k(\theta),
\end{equation}
where $F_k$ is the local loss function of client $k$ with model parameters $\theta$, and $\rho_k \geq 0$ with $\sum_{k=1}^{K} \rho_k=1$ specifies the relative weight of client $k$. In practice, $F_k$ is typically defined as the empirical risk over client $k$'s local dataset $\{(x^{(k)}_i, y^{(k)}_i)\}_{i=1}^{N^{(k)}}$ of size $N^{(k)}$, i.e., $F_k(\theta)={1}/{N^{(k)}} \sum_{i=1}^{N^{(k)}} \ell\bigl(x^{(k)}_i, y^{(k)}_i;\theta\bigr)$, where $\ell$ denotes the loss for a given training sample.
Common choices include the $\ell_2$ loss and the cross-entropy loss, see \cite{calin2020deep} for more options.
The relative weight $\rho_k$ is often chosen as $\rho_k=N^{(k)} / N_K$, where $N_K = \sum_{k=1}^{K}N^{(k)}$ is the total size of all clients' datasets.

\subsection{FedAvg Algorithm}
\label{subsec_pre_FL}

To solve \eqref{eq_FL}, FL algorithms normally involve a sequence of training rounds in which the clients perform local updates and the server performs aggregation. To fix ideas, we focus on FedAvg \cite{mcmahan2017communicationefficient}, which is the most commonly used algorithm in FL.

In FedAvg, at each round $t$, the server selects a subset $\mathcal{K}_t$ of clients to participate and sends them the current global model parameters $\theta_t$.
Each selected client $k \in \mathcal{K}_t$ then sets its local model parameters $\theta_{t}^{(k)} = \theta_t$ and updates $\theta_{t}^{(k)}$ for $E$ epochs, each consisting of $B$ mini-batches.
After training, client $k$'s model update $\Delta\theta_t^{(k)}$ can be written as
\begin{equation}\label{eq_delta_theta}
	\Delta\theta_t^{(k)} = - \eta \sum_{e=1}^E \sum_{b=1}^{B} \nabla_{\theta_{t,e,b}^{(k)}} \ell\left(X_{t,e,b}^{(k)}, Y_{t,e,b}^{(k)}\right),
\end{equation}
where $\ell(X_{t,e,b}^{(k)}, Y_{t,e,b}^{(k)})$ denotes the loss evaluated on the mini-batch $\big(X_{t,e,b}^{(k)}, Y_{t,e,b}^{(k)}\big)$ with respect to the model parameters $\theta_{t,e,b}^{(k)}$. The subscripts $t$, $e$, and $b$ index the round, epoch, and mini-batch, respectively. Consequently, client $k$'s local model parameters are updated to
\begin{equation}\label{eq_model_update_mini_batch}
	\theta_{t+1}^{(k)} = \theta_{t}^{(k)} + \Delta\theta_t^{(k)}.
\end{equation}
Finally, client $k$ sends its updated local model parameters $\theta_{t+1}^{(k)}$ back to the server for averaging.

\begin{figure*}[t]
	\centering
	\includegraphics[width=0.99\textwidth]{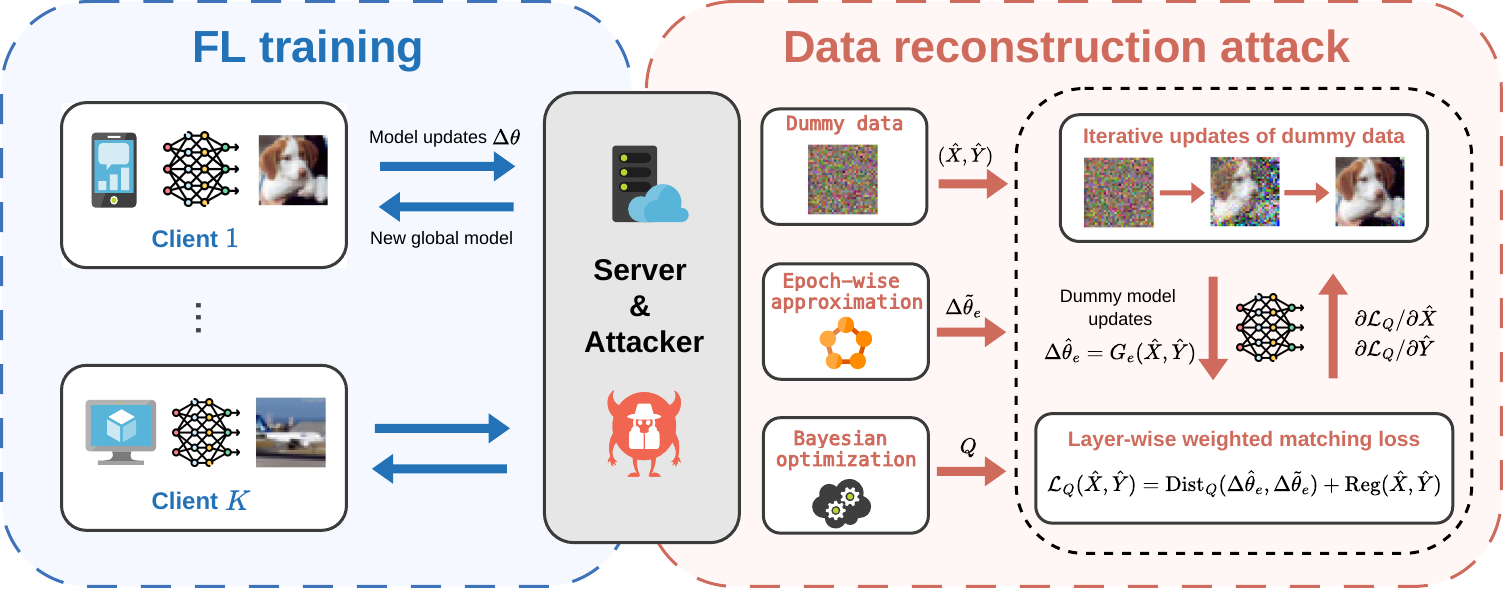}
	\caption{Illustration of the FL training process and the approximate and weighted attack process.}
	\label{fig_attack_process}
\end{figure*}

\subsection{Vanilla Data Reconstruction Attack}
\label{subsec_DRA}
Although clients share only updated model parameters with the server, their private training data remain vulnerable to data reconstruction attacks.
In this subsection, we introduce the formulation and general procedure of a data reconstruction attack.

For notational simplicity, we consider an arbitrary attack on client $k$ at round $t$. We then omit the client and round indices $(k,t)$ whenever no ambiguity arises.

As shown in \eqref{eq_delta_theta}, during the local training process, the model update $\Delta\theta$ consists of gradients computed over $E \times B$ mini-batches. Let $G$ denote the map from the training data $(X, Y)$ to the model update $\Delta\theta$. Then \eqref{eq_delta_theta} can be written compactly as
\begin{equation}\label{eq_delta_theta_G}
	\Delta\theta = G\left(X, Y\right).
\end{equation}

For an attacker with access to $\Delta\theta$, reconstructing $(X, Y)$ is essentially an inverse problem. In particular, if $G^{-1}$ exists and is known analytically, the attacker can recover $(X, Y)$ directly as follows:
\begin{equation}\label{eq_inverse_G}
	\left(X, Y\right) = G^{-1}\left(\Delta\theta\right).
\end{equation}
However, since neural networks are highly nonlinear and the model updates are aggregated over multiple mini-batches, it is generally difficult to identify $G^{-1}$. To address this, we introduce a numerical approach for solving \eqref{eq_delta_theta_G}.

Assuming that the attacker knows the client's training procedure and can therefore evaluate $G$, problem \eqref{eq_delta_theta_G} can be approached numerically as follows.
First, to launch the attack, the attacker randomly initializes a dummy dataset $(\hat{X}, \hat{Y})$ with the same dimensions as the client's ground-truth dataset $(X, Y)$. The attacker then uses $G$ to calculate the dummy model update $\Delta\hat{\theta}$ given by
\begin{equation}\label{eq_delta_theta_hat_G}
	\Delta\hat{\theta} = G(\hat{X}, \hat{Y}).
\end{equation}
Then, the attacker reconstructs the client's data by minimizing the following loss function:
\begin{equation}\label{eq_loss_grad_match}
\mathcal{L}_{m}(\hat{X}, \hat{Y}) = \operatorname{Dist}(\Delta\hat{\theta}, \Delta\theta) + \operatorname{Reg}(\hat{X}, \hat{Y}).
\end{equation}
Here, $\operatorname{Reg}$ denotes regularization terms that improve reconstruction quality, such as total variation regularization \cite{geiping2020inverting} or batch normalization regularization \cite{yin2021see}. The $\operatorname{Dist}$ term quantifies the discrepancy between the dummy and true model updates. Common choices include the $\ell_2$ distance \cite{zhao2020idlg} and cosine distance \cite{xu2022agic}, with the latter defined as
\begin{equation}\label{eq_loss_cosine}
	\| \Delta\hat{\theta} - \Delta\theta \|_{\text{cos}} = 1 - \frac{\langle \Delta\hat{\theta}, \Delta\theta \rangle}{\|\Delta\hat{\theta}\|\,\|\Delta\theta\|}.
\end{equation}%
Finally, the reconstructed data $(\hat{X}^*, \hat{Y}^*)$ can be obtained by solving the following optimization problem:
\begin{equation}\label{eq_inv_arg_min}
	(\hat{X}^*, \hat{Y}^*) = \arg \min_{\hat{X}, \hat{Y}} \mathcal{L}_{m}(\hat{X}, \hat{Y}).
\end{equation}
The iterative updates of $\hat{X}$ and $\hat{Y}$ can be carried out using gradient descent with learning rate $\hat{\eta}$:
\begin{equation*}
	\begin{aligned}
	\hat{X}^{+} &= \hat{X} - \hat{\eta} \nabla_{\hat{X}}
	\mathcal{L}_{m}(\hat{X}, \hat{Y}),\\
	\hat{Y}^{+} &= \hat{Y} - \hat{\eta} \nabla_{\hat{Y}}
	\mathcal{L}_{m}(\hat{X}, \hat{Y}).
	\end{aligned}
\end{equation*}

\begin{remark}[An honest-but-curious server (attacker)]
\label{rem_attacker}
	A data reconstruction attacker needs to evaluate the mapping $G$ in \eqref{eq_delta_theta_hat_G}, which encapsulates the client's local training procedure. Without knowledge of the local hyperparameters, reproducing $G$, or equivalently replicating the client's local training, is generally infeasible.
	However, in practice, the central server holds substantial information about the training process. Consequently, an honest-but-curious server can conduct data reconstruction attacks when it has access to the following information:
	\begin{enumerate}
		\item {Model architecture}: The server normally determines the neural network architecture shared by all clients.
		\item {Initial model parameters}: The client's initial local model parameters $\theta$ are dispatched by the server.
		\item {Model update}: Each client sends the updated model parameters $\theta^{+}$ back to the server. Thus, the server can readily obtain the model update $\Delta\theta = \theta^{+} - \theta$.
		\item {Loss function}: The server typically also knows the form of the loss function $\ell$ shared by all clients. This choice usually remains unchanged during training.
		\item {Dataset size}: This information is shared with the server for weighted aggregation.
		\item {Client's local training hyperparameters}: The server can assign hyperparameters such as the number of epochs $E$, the number of mini-batches $B$, and the learning rate $\eta$ to control the training workload and procedure.
	\end{enumerate}
\end{remark}

However, unless additional leakage is assumed, the attacker does \emph{not} observe the client's random reshuffling of the dataset at each epoch, the resulting mini-batch partitions, or the intermediate local model parameters. Therefore, the attacker's knowledge is fixed, and the differences among the four scenarios discussed below lie in whether the information visible to the server is sufficient to evaluate $G$ exactly.

\section{Analysis of Data Reconstruction under Different FedAvg Settings}
\label{sec_different_EB}

In this section, we compare four FedAvg scenarios determined by the number of local epochs $E$ and the number of mini-batches $B$. The adversary is always the same honest-but-curious server (attacker) described in \Cref{rem_attacker}. Hence, these four scenarios are not distinguished by different attacker capabilities, but rather by whether the information available to the attacker is sufficient to evaluate the client's local training map $G$ in \eqref{eq_delta_theta_hat_G} and thus calculate the dummy model update.

\par\smallskip
\noindent\textbf{Scenario 1: $E = 1, B = 1$.} In this scenario, the attacker can evaluate $G$ exactly and replicate the client's training process.
Indeed, since $E = 1$ and $B = 1$, the client uses full-batch gradient descent for one epoch as follows:
\begin{equation}
	\begin{aligned}
		G(X, Y)
		&= - \eta \nabla_{\theta} \ell\left(X, Y\right).
	\end{aligned}
\end{equation}
Since the update is computed from a single full batch at the known parameter $\theta$, no hidden shuffling or intermediate state is involved. Thus, the attacker can replicate the client's training process by replacing $(X, Y)$ with the dummy dataset $(\hat{X}, \hat{Y})$.

\par\smallskip
\noindent\textbf{Scenario 2: $E > 1, B = 1$.} The attacker can also evaluate $G$ exactly in this scenario. Specifically, the client uses full-batch gradient descent for $E$ epochs:
\begin{equation}
	G(X, Y) = - \eta \sum_{e=1}^E \nabla_{\theta_{e}} \ell\left(X, Y\right).
\end{equation}
Although there are multiple epochs, every epoch still uses the whole dataset since $B = 1$. Hence, there is no hidden mini-batch partition to infer, and the attacker can roll out exactly the same sequence of intermediate parameters when replacing $(X, Y)$ with the dummy dataset $(\hat{X}, \hat{Y})$ and training the model for $E$ epochs. Therefore, Scenario 2 remains exactly reproducible.

\par\smallskip
\noindent\textbf{Scenario 3: $E = 1, B > 1$.}
In this scenario, since $B > 1$, the client uses mini-batch gradient descent for one epoch in the following way:
\begin{equation}\label{eq_scenario_3}
	G(X, Y) = - \eta \sum_{b=1}^B \nabla_{\theta_{b}} \ell\left(X_{b}, Y_{b}\right).
\end{equation}
The key point is that the hidden randomness occurs only once because $E = 1$. Consequently, the dataset is shuffled only once. The attacker can first partition its dummy dataset $(\hat{X}, \hat{Y})$ into $B$ mini-batches $\{(\hat{X}_{b}, \hat{Y}_{b})\}_{b=1}^B$. Then, by replacing $(X_{b}, Y_{b})$ in \eqref{eq_scenario_3} with $(\hat{X}_{b}, \hat{Y}_{b})$, it can reproduce the client's training process. Therefore, Scenario 3 remains tractable for existing reconstruction attacks.

\par\smallskip
\noindent\textbf{Scenario 4: $E > 1, B > 1$.} In this scenario, the attacker cannot evaluate the exact map $G$ needed to calculate the dummy model update $\Delta\hat{\theta}$ defined in \eqref{eq_delta_theta_hat_G}.
More concretely, the client uses mini-batch gradient descent for $E$ epochs:
\begin{equation}\label{eq_scenario4}
	G(X, Y) = - \eta \sum_{e=1}^E \sum_{b=1}^B \nabla_{\theta_{e,b}} \ell\left(X_{e,b}, Y_{e,b}\right).
\end{equation}
In each epoch, the client first reshuffles its dataset and then partitions it into $B$ mini-batches. Consequently, the final update depends on a sequence of hidden mini-batch partitions across epochs. However, information about this epoch-wise shuffling is never available to the attacker. If the attacker simply performs $E$ epochs of updates with arbitrarily chosen mini-batch orders, the resulting dummy training is no longer guaranteed to match the client's original training process.

Overall, while existing attack methods are generally applicable to Scenarios 1--3, the substantially more challenging Scenario 4 has received limited attention. To overcome the difficulties inherent to this scenario, we propose an interpolation-based approximation method. By interpolating between the initial and the final updated model parameters, the attacker can effectively estimate the hidden intermediate model updates for each epoch. This approximation reduces the intractable Scenario 4 to independent instances of the tractable Scenario 3. A detailed formulation of the proposed method is presented in the subsequent section.

\section{Approximate and Weighted Attack in Data Reconstruction}
\label{sec_method}

In this section, we propose an approximate and weighted data reconstruction attack. The algorithm is summarized in \Cref{alg_DataRecAttack} and illustrated in \Cref{fig_attack_process}.

\subsection{Approximation of the Intermediate Model Updates}
\label{sec_approximate_G}

As discussed in \Cref{sec_different_EB}, when $E > 1$ and $B > 1$, reshuffling the dataset at each epoch makes it difficult for the attacker to reproduce $G$. However, if the attacker knew the intermediate model update $\Delta\theta_{e}$ for any epoch $e \in \{1,2, \ldots, E\}$, the problem could be reduced from Scenario 4 ($E > 1$ and $B > 1$) to Scenario 3 ($E = 1$ and $B > 1$) by attacking each epoch separately.

To this end, we interpolate between the updated model parameters $\theta^{+}$ and the initial model parameters $\theta$ to approximate the intermediate model parameters at the end of each epoch.
In particular, for a client that trains its model for $E$ epochs, the approximate intermediate model parameters $\{\tilde \theta_{e}\}_{e=1}^{E}$ are given by
\begin{equation}\label{eq_theta_approximate}
	\tilde{\theta}_{e} = \theta + \frac{e}{E}\left(\theta^{+} - \theta\right), \quad e = 1, 2, \ldots, E.
\end{equation}
Then, the approximate model updates $\{\Delta\tilde{\theta}_{e}\}_{e=1}^{E}$ corresponding to each epoch can be obtained as
\begin{equation}\label{eq_model_update_approximate}
	\Delta\tilde{\theta}_{e} = \tilde{\theta}_{e} - \tilde{\theta}_{e-1}, \quad e = 1, 2, \ldots, E,
\end{equation}
where $\tilde{\theta}_{0} = \theta$ is the initial model parameters of the round.

After obtaining the approximate updates $\{\Delta\tilde{\theta}_{e}\}_{e=1}^{E}$, the attacker can use them as surrogates for the unknown intermediate model updates $\Delta\theta_{e}$. In other words, the attack problem is reduced from Scenario 4 ($E > 1$ and $B > 1$) to a one-epoch Scenario 3 problem ($E = 1$ and $B > 1$).

We denote by $G_{e}$ the client's training process at epoch $e$. Then, the dummy model update $\Delta\hat{\theta}_{e}$ can be obtained according to \eqref{eq_scenario_3} as
\begin{equation}\label{eq_scenario3_2}
	\Delta\hat{\theta}_{e} = G_{e}(\hat{X}, \hat{Y}) = - \eta \sum_{b=1}^B \nabla_{\theta_{e,b}} \ell\left(\hat{X}_{e,b}, \hat{Y}_{e,b}\right). 
\end{equation}
Finally, the attack can be carried out by following the procedure of Scenario 3. The same idea also applies when each epoch contains only one mini-batch. By leveraging interpolation, the attacker can reconstruct the data by training for a single epoch using the approximate model update, which substantially reduces computational overhead, especially when $E$ is large.
Under linear interpolation, the approximate epoch-wise update $\Delta\tilde{\theta}_{e}$ is identical for all $e$. This approximation is reasonable, particularly when training involves many epochs and a small learning rate. Its effectiveness is demonstrated through extensive experiments in \Cref{sec_simulation}. Other interpolation schemes may also be considered and deserve further investigation.

\subsection{Improved Weighted Loss Function for the Data Reconstruction Attack}
\label{sec_weighted_loss_function}

The commonly used loss function \eqref{eq_loss_grad_match} for data reconstruction treats different components of the model update $\Delta\tilde{\theta}_{e}$ equally.
However, as observed in \cite{chen2019shallowing}, different layers in a neural network contribute differently to performance. Motivated by this observation, we assign different weights to different layers to facilitate reconstruction. The construction of our method is described below.

Consider a neural network with $L$ layers. The model update $\Delta\tilde{\theta}_{e}$ can be decomposed into layer-wise updates as $\Delta\tilde{\theta}_{e} = \{\Delta\tilde{\theta}_{e}^{(l)}\}_{l=1}^L$. Similarly, the dummy model update can be written as $\Delta\hat{\theta}_{e} = \{\Delta\hat{\theta}_{e}^{(l)}\}_{l=1}^L$.
By assigning a positive weight $q^{(l)}$ to layer $l$, the loss function in \eqref{eq_loss_grad_match} is modified as
\begin{equation}\label{eq_weighted_loss}
	\mathcal{L}_{Q}(\hat{X}, \hat{Y}) = \operatorname{Dist}_{Q}(\Delta\hat{\theta}_{e}, \Delta\tilde{\theta}_{e}) + \operatorname{Reg}(\hat{X}, \hat{Y}).
\end{equation}
Here, when cosine distance is used, $\operatorname{Dist}_{Q}$ is defined as
\begin{equation}\label{eq_weighted_cosine_distance}
1-\frac{\sum_{l=1}^{L} q^{(l)}\langle \Delta\hat{\theta}_{e}^{(l)}, \Delta\tilde{\theta}_{e}^{(l)} \rangle}
{\sqrt{\sum_{l=1}^{L} q^{(l)}\|\Delta\hat{\theta}_{e}^{(l)}\|_2^2}
\sqrt{\sum_{l=1}^{L} q^{(l)}\|\Delta\tilde{\theta}_{e}^{(l)}\|_2^2}}.
\end{equation}

The weighted loss function leverages the distinct characteristics of different layers in the model update. In the next subsection, we introduce a systematic method for designing the layer weights to enhance reconstruction performance.

\subsubsection{Design of Layer Weights}
\par\smallskip
\noindent\textbf{Increasing weights layer by layer.}
We consider linearly increasing weight profiles for different types of layers. To illustrate the idea clearly, we focus on the commonly used ResNet architecture \cite{he2016deep}, which contains convolutional, batch normalization, and fully connected layers. Specifically, we design the following weight functions for each type of layer:
\begin{subequations}\label{eq_q_max_3kinds}
	\begin{align}
		\begin{split}\label{eq_q_max_conv}
			q_{\mathrm{cv}}^{(l)} &=
			\begin{cases}
				\frac{q_{\mathrm{cv}}-1}{L_{\mathrm{cv}}-1}(l-1) + 1, & 1 \leq l \leq L_{\mathrm{cv}},\ L_{\mathrm{cv}} > 1, \\
				q_{\mathrm{cv}}, & l = L_{\mathrm{cv}} = 1,
			\end{cases}
		\end{split}\\
		\begin{split}\label{eq_q_max_bn}
			q_{\mathrm{bn}}^{(l)} &=
			\begin{cases}
				\frac{q_{\mathrm{bn}}-1}{L_{\mathrm{bn}}-1}(l-1) + 1, & 1 \leq l \leq L_{\mathrm{bn}},\ L_{\mathrm{bn}} > 1, \\
				q_{\mathrm{bn}}, & l = L_{\mathrm{bn}} = 1,
			\end{cases}
		\end{split}\\
		\begin{split}\label{eq_q_max_fc}
			q_{\mathrm{fc}}^{(l)} &=
			\begin{cases}
				\frac{q_{\mathrm{fc}}-1}{L_{\mathrm{fc}}-1}(l-1) + 1, & 1 \leq l \leq L_{\mathrm{fc}},\ L_{\mathrm{fc}} > 1, \\
				q_{\mathrm{fc}}, & l = L_{\mathrm{fc}} = 1,
			\end{cases}
		\end{split}
	\end{align}
\end{subequations}
where $L_{\mathrm{cv}}$, $L_{\mathrm{bn}}$, and $L_{\mathrm{fc}}$ denote the numbers of convolutional, batch normalization, and fully connected layers, respectively, and $L = L_{\mathrm{cv}} + L_{\mathrm{bn}} + L_{\mathrm{fc}}$. The values $q_{\mathrm{cv}}>1$, $q_{\mathrm{bn}}>1$, and $q_{\mathrm{fc}}>1$ are the maximum weights assigned to the last layer of each respective type. For a given neural network with a fixed number of layers ($L_{\mathrm{cv}}$, $L_{\mathrm{bn}}$, and $L_{\mathrm{fc}}$), the values of $q_{\mathrm{cv}}$, $q_{\mathrm{bn}}$, and $q_{\mathrm{fc}}$ determine the slopes of the linearly increasing weight functions in \eqref{eq_q_max_3kinds}.

\par\smallskip
\noindent\textbf{Enhancing the weights of layers with larger errors.}
Applying the linearly increasing weights determined by \eqref{eq_q_max_3kinds} in the loss function \eqref{eq_weighted_loss} may overemphasize certain layers and lead to biased reconstructions. To balance the benefits of linearly increasing weights against this risk, we modify the weights of layers with larger errors by exploiting statistical information, namely the mean $\mu(\cdot)$ and variance $\sigma^2(\cdot)$ of the layer-wise model updates. The enhancement procedure is described below.

First, we compute the relative errors $e_{\mathrm{mean}}^{(l)}$ and $e_{\mathrm{var}}^{(l)}$ between the dummy model update $\Delta\hat{\theta}_{e}^{(l)}$ and the target approximate model update $\Delta\tilde{\theta}_{e}^{(l)}$ at each layer as follows:
\begin{equation}\label{eq_error_mean}
	e_{\mathrm{mean}}^{(l)} = \frac{|\mu(\Delta\hat{\theta}_{e}^{(l)}) - \mu(\Delta\tilde{\theta}_{e}^{(l)})|}{|\mu(\Delta\tilde{\theta}_{e}^{(l)})|}, \quad l = 1, 2, \ldots, L,
\end{equation}
\begin{equation}\label{eq_error_var}
	e_{\mathrm{var}}^{(l)} = \frac{|\sigma^2(\Delta\hat{\theta}_{e}^{(l)}) - \sigma^2(\Delta\tilde{\theta}_{e}^{(l)})|}{|\sigma^2(\Delta\tilde{\theta}_{e}^{(l)})|}, \quad l = 1, 2, \ldots, L.
\end{equation}

Next, we select a subset $\mathcal{P} \subseteq [L] \mathrel{:=} \{1, 2, \ldots, L\}$ of layers with the largest relative errors according to $\{e_{\mathrm{mean}}^{(l)}\}_{l=1}^L$ and $\{e_{\mathrm{var}}^{(l)}\}_{l=1}^L$. For these layers, we replace their original weights with a fixed value $q_{\mathrm{en}}$, by setting
\begin{equation}\label{eq_weight_aug}
	q^{(l)} = q_{\mathrm{en}}, \quad l \in \mathcal{P}.
\end{equation}
Here, $q_{\mathrm{en}} > 1$ is a hyperparameter to be determined. This replacement strategy, rather than multiplying the existing weights by another constant, prevents assigning excessively large weights in deeper layers, which could lead to instability or degradation of reconstruction performance.

The choice of the subset $\mathcal{P}$ is determined by the proportional parameters $p_{\mathrm{mean}} \in [0, 1]$ and $p_{\mathrm{var}} \in [0, 1]$. For a given $p_{\mathrm{mean}}$, we first select $N_{\mathrm{mean}} = \lceil p_{\mathrm{mean}} \cdot L \rceil$ layers with the largest relative errors in $\{e_{\mathrm{mean}}^{(l)}\}_{l=1}^L$.
Let the set of indices corresponding to the $N_{\mathrm{mean}}$ layers be denoted as $\mathcal{P}_{\mathrm{mean}}$, that is
\begin{equation}
	\mathcal{P}_{\mathrm{mean}} = \{i_1, i_2, \ldots, i_{N_{\mathrm{mean}}}\},
\end{equation}
where $i_1, i_2, \ldots, i_{N_{\mathrm{mean}}} \in [L]$ are selected such that $e_{\mathrm{mean}}^{(i_1)} \geq e_{\mathrm{mean}}^{(i_2)} \geq \ldots \geq e_{\mathrm{mean}}^{(i_{N_{\mathrm{mean}}})} \geq e_{\mathrm{mean}}^{(l)}$ for all $l \in [L] \setminus \mathcal{P}_{\mathrm{mean}}$.

Similarly, we obtain a set $\mathcal{P}_{\mathrm{var}}$ with $N_{\mathrm{var}} = \lceil p_{\mathrm{var}} \cdot L \rceil$ elements as
\begin{equation}
	\mathcal{P}_{\mathrm{var}} = \{j_1, j_2, \ldots, j_{N_{\mathrm{var}}}\},
\end{equation}
where $j_1, j_2, \ldots, j_{N_{\mathrm{var}}} \in [L]$ are selected such that $e_{\mathrm{var}}^{(j_1)} \geq e_{\mathrm{var}}^{(j_2)} \geq \ldots \geq e_{\mathrm{var}}^{(j_{N_{\mathrm{var}}})} \geq e_{\mathrm{var}}^{(l)}$ for all $l \in [L] \setminus \mathcal{P}_{\mathrm{var}}$.

Finally, the subset $\mathcal{P}$ can be obtained as the intersection of $\mathcal{P}_{\mathrm{mean}}$ and $\mathcal{P}_{\mathrm{var}}$:
\begin{equation}\label{eq_subsec_P}
	\mathcal{P} = \mathcal{P}_{\mathrm{mean}} \cap \mathcal{P}_{\mathrm{var}}.
\end{equation}

\par\smallskip
\noindent\textbf{Hyperparameters to tune.}
Following \eqref{eq_q_max_3kinds} and \eqref{eq_weight_aug}, the layer weights $\{q^{(l)}\}_{l=1}^L$ in \eqref{eq_weighted_loss} are determined by the parameter vector $Q\in \mathbb{R}^6$, defined as
\begin{equation}\label{eq_Q}
	Q = (q_{\mathrm{cv}}, q_{\mathrm{bn}}, q_{\mathrm{fc}}, q_{\mathrm{en}}, p_{\mathrm{mean}}, p_{\mathrm{var}}).
\end{equation}
Given $Q$, the reconstructed data $(\hat{X}^*, \hat{Y}^*)$ are obtained by solving the following optimization problem:
\begin{equation}\label{eq_argmin_weighted_loss}
	(\hat{X}^*, \hat{Y}^*) = \arg \min_{\hat{X}, \hat{Y}} \mathcal{L}_{Q}(\hat{X}, \hat{Y}).
\end{equation}
We then use Bayesian optimization to choose an appropriate $Q$ for better reconstruction.

\subsubsection{Choice of $Q$ by Bayesian Optimization}\label{sec_BO}
\par\smallskip
\noindent\textbf{Objective function.}
As shown in \eqref{eq_argmin_weighted_loss}, for a given $Q$ one can obtain the reconstructed data $(\hat{X}^*, \hat{Y}^*)$. Then, the corresponding reconstructed model update can be computed as $\Delta\hat{\theta}_{e}^* = G_{e}(\hat{X}^*, \hat{Y}^*)$.

Let $f: \mathbb{R}^6 \rightarrow \mathbb{R}$ be the objective function that quantifies the discrepancy between the reconstructed model update $\Delta\hat{\theta}_{e}^*$ and the target approximate model update $\Delta\tilde{\theta}_{e}$. Specifically, we use cosine distance as the metric:
\begin{equation}\label{eq_obj_func_BO_f_Q}
	f(Q) = \| \Delta\hat{\theta}_{e}^* - \Delta\tilde{\theta}_{e} \|_{\text{cos}}.
\end{equation}
Finding the optimal $Q^*$ is equivalent to solving the following optimization problem:
\begin{equation}\label{eq_argmin_Q}
	Q^* = \arg \min_{Q} f(Q).
\end{equation}
For this optimization problem, $f$ is a black-box function that does not admit an analytic expression. Moreover, evaluating $f$ is computationally expensive because each evaluation requires a full data reconstruction attack to obtain $\Delta\hat{\theta}_{e}^*$. As a result, traditional parameter-selection methods such as grid search are not feasible. To overcome these difficulties, we employ Bayesian optimization \cite{frazier2018tutorial} to solve \eqref{eq_argmin_Q}.

\par\smallskip
\noindent\textbf{A Bayesian optimization algorithm for \eqref{eq_argmin_Q}.}
Bayesian optimization is a powerful technique for optimizing black-box functions that are expensive to evaluate and may involve noise or other sources of uncertainty. In general, Bayesian optimization iteratively uses a surrogate model to approximate the black-box function and then employs an acquisition function to determine the next set of parameters to evaluate.

The surrogate model approximates the black-box objective function $f$ and is commonly chosen to be a Gaussian process (GP) \cite{rasmussen2006gaussian}. Formally, a GP is a collection of random variables, any finite number of which have a joint Gaussian distribution.

Given an initial set $\mathcal{O} = \left\{(Q_i, f(Q_i))\right\}_{i=1}^n$ that contains $n$ sampled points and their function values, the resulting prior distribution is
\begin{equation}\label{eq_GP_prior}
	\boldsymbol{f} \sim \mathcal{N}\left(\boldsymbol{\mu}, \Sigma_{\boldsymbol{Q} \boldsymbol{Q}}\right),
\end{equation}
where $\boldsymbol{f} = \left(f(Q_1), \ldots, f(Q_n)\right)^\top$, $\boldsymbol{Q} = \left(Q_1, \ldots, Q_n\right)^\top$, and $\boldsymbol{\mu}=\left(\mu\left(Q_1\right), \ldots, \mu\left(Q_n\right)\right)^\top$, with $\mu(\cdot)$ denoting the mean function, which is commonly set to zero. Moreover, $\Sigma_{\boldsymbol{Q} \boldsymbol{Q}} \in \mathbb{R}^{n \times n}$ is the matrix of covariances with its $(i, j)$ entry given by $\kappa\left(Q_i, Q_j\right)$, and $\kappa(\cdot, \cdot)$ is a positive definite kernel function, typically chosen as the Gaussian kernel.

Then, we can infer the value of $f(Q)$ at a new point $Q$ by computing the posterior distribution of $f(Q)$ given prior observations \cite{rasmussen2006gaussian} as follows:
\begin{equation}\label{eq_GP_posterior}
	\begin{aligned}
		f(Q) \mid \boldsymbol{f} &\sim \mathcal{N}\left(\mu_Q, \sigma^2_Q\right), \\
		\mu_Q & = \Sigma_{Q \boldsymbol{Q}} \Sigma_{\boldsymbol{Q} \boldsymbol{Q}}^{-1} \boldsymbol{f}, \\
		\sigma^2_Q & = \Sigma_{Q Q} - \Sigma_{Q \boldsymbol{Q}} \Sigma_{\boldsymbol{Q} \boldsymbol{Q}}^{-1} \Sigma_{\boldsymbol{Q} Q}.
	\end{aligned}
\end{equation}

The acquisition function is used to propose the parameter vector for the next trial by balancing exploitation and exploration. Exploitation means sampling at locations where the surrogate model predicts a low objective value, whereas exploration means sampling at locations where the predictive uncertainty is high.

One of the most popular acquisition functions is expected improvement (EI) \cite{mockus1975bayesian}. Let $f_{\min}$ denote the best function value obtained so far. Then, the improvement over $f_{\min}$ at point $Q$ can be defined as
\begin{equation}
	I(Q) = \max (f_{\min} - f(Q), 0).
\end{equation}
The improvement $I(Q)$ is a random variable since $f(Q) \sim \mathcal{N}\left(\mu_Q, \sigma^2_Q\right)$ as shown in \eqref{eq_GP_posterior}. To obtain the expected improvement, we can take the expected value as follows:
\begin{equation}
	EI(Q)=\mathbb{E}[\max (f_{\min} - f(Q), 0)].
\end{equation}
The expected improvement can be evaluated analytically under the GP \cite{jones1998efficient} and is given by
\begin{equation}\label{eq_EI}
	\begin{aligned}
		EI(Q) =& \left(f_{\min}-\mu_Q\right) \Phi\left(\frac{f_{\min}-\mu_Q}{\sigma_Q}\right)\\
		&+\sigma_Q \varphi\left(\frac{f_{\min}-\mu_Q}{\sigma_Q}\right),
	\end{aligned}
\end{equation}
where $\varphi$ and $\Phi$ are the probability density and cumulative distribution functions of the standard normal distribution, respectively. Thus, $EI(Q)$ is higher for a point $Q$ predicted to have a smaller $\mu_Q$ and a larger $\sigma_Q$, indicating the trade-off between exploitation and exploration.

Given the $EI$, the parameter $Q_{n+1}$ for the next trial is chosen to be the one with the largest expected improvement:
\begin{equation}\label{eq_argmax_EI}
	Q_{n+1} = \arg\max_{Q} EI(Q).
\end{equation}
Evaluating $EI(Q)$ is much easier than evaluating the function $f$ in \eqref{eq_argmin_Q}. The optimization problem \eqref{eq_argmax_EI} can be solved using classical techniques such as Newton's method.

Finally, based on the above discussion, we summarize our approximate and weighted data reconstruction attack method in \Cref{alg_DataRecAttack}.

\begin{algorithm}[!h]
	\caption{Approximate and Weighted Attack (AWA).}
	\label{alg_DataRecAttack}
	\begin{algorithmic}[1]
		\STATE{Intercept a client's model update $\Delta\theta$.}
		\STATE{Approximate the epoch-wise $\Delta \tilde \theta_{e}$ and $G_{e}$. $\quad \triangleright$ \eqref{eq_model_update_approximate}}
		\STATE{Initialize an empty set $\mathcal{O}$.}
		\FOR{each trial $i = 1, 2, \ldots, n$}
		\STATE{Generate a random test point $Q_i$.}
		\STATE{Obtain $\hat{X}$, $\hat{Y}$, and $f(Q_i)$ via $\textbf{RecAttack}(Q_i, \Delta\tilde{\theta}_{e}, G_{e})$.}
		\STATE{Update $\mathcal{O} \leftarrow \mathcal{O} \cup \{(Q_i, f(Q_i))\}$.}
		\ENDFOR
		\FOR{each trial $i = n+1, n+2, \ldots, N_{\mathrm{BO}}$}
		\STATE{Fit a GP of $f$ with the samples in $\mathcal{O}$. $\quad \triangleright$ \eqref{eq_GP_prior}}
		\STATE{Choose $Q_i$ with the largest EI. $\quad \triangleright$ \eqref{eq_argmax_EI}}
		\STATE{Obtain $\hat{X}$, $\hat{Y}$, and $f(Q_i)$ via $\textbf{RecAttack}(Q_i, \Delta\tilde{\theta}_{e}, G_{e})$.}
		\STATE{Update $\mathcal{O} \leftarrow \mathcal{O} \cup \{(Q_i, f(Q_i))\}$.}
		\ENDFOR
		\STATE{Set $Q^* = \arg \min_{Q_i} \{f(Q_i)\}_{i=1}^{N_{\mathrm{BO}}}$.}
		\STATE{Obtain $\hat{X}^*$, $\hat{Y}^*$ via $\textbf{RecAttack}(Q^*, \Delta\tilde{\theta}_{e}, G_{e})$.}
		\RETURN{$\hat{X}^*$, $\hat{Y}^*$.}
	\end{algorithmic}
	\begin{algorithmic}
		\STATE
	\end{algorithmic}
	\begin{algorithmic}[1]
		\STATE{$\textbf{RecAttack}(Q_i, \Delta\tilde{\theta}_{e}, G_{e}):$}
		\STATE{Initialize the dummy data $(\hat{X}, \hat{Y})$ and set $Q = Q_i$.}
		\FOR{each attack iteration from $1$ to $N_{\mathrm{AT}}$}
		\STATE{Compute $\Delta\hat{\theta}_{e} = G_{e}(\hat{X}, \hat{Y})$.}
		\STATE{Calculate $\{q^{(l)}\}_{l=1}^L$ based on $Q$. $\quad \triangleright$ \eqref{eq_q_max_3kinds}}
		\STATE{Select layers $l \in \mathcal{P}$ with the largest errors. $\quad \triangleright$ \eqref{eq_subsec_P}}
		\STATE{Set $q^{(l)} = q_{\mathrm{en}}$ for all $l \in \mathcal{P}$ based on $Q$. $\quad \triangleright$ \eqref{eq_weight_aug}}
		\STATE{Calculate the weighted loss $\mathcal{L}_{Q}(\hat{X}, \hat{Y})$. $\quad \triangleright$ \eqref{eq_weighted_loss}}
		\STATE{Update $\hat{X}^{+} = \hat{X} - \hat{\eta} \nabla_{\hat{X}} \mathcal{L}_{Q}(\hat{X}, \hat{Y})$.}
		\STATE{Update $\hat{Y}^{+} = \hat{Y} - \hat{\eta} \nabla_{\hat{Y}} \mathcal{L}_{Q}(\hat{X}, \hat{Y})$.}
		\STATE{Set $\hat{X} = \hat{X}^{+}$ and $\hat{Y} = \hat{Y}^{+}$.}
		\ENDFOR
		\STATE{Calculate $f(Q) = \| G_{e}(\hat{X}, \hat{Y}) - \Delta\tilde{\theta}_{e} \|_{\text{cos}}$.}
		\RETURN{$\hat{X}$, $\hat{Y}$, $f(Q)$.}
	\end{algorithmic}
\end{algorithm}

\section{Numerical Experiments}
\label{sec_simulation}
In this section, we report numerical results. We first describe the experimental setup and implementation details. We then explain the choice of hyperparameters and evaluation metrics. Finally, we test AWA on image data reconstruction and compare its performance with three state-of-the-art methods, namely iDLG \cite{zhao2020idlg}, AGIC \cite{xu2022agic}, and GENG \cite{geng2023improved}, across various scenarios. 
The source code for our AWA method is available at \url{https://github.com/DCN-FAU-AvH/FL-AWA-attack}.

\subsection{Experimental Setup}

\noindent\textbf{Hardware.} All experiments were conducted on a cluster node at FAU equipped with AMD EPYC 7662 CPUs and NVIDIA A100/V100 GPUs.

\par\smallskip
\noindent\textbf{Implementation details.}
In our implementation, we use images from ImageNet \cite{deng2009imagenet} (color images from 1,000 categories, each of size $224\times224$) as the clients' ground-truth training data. The model used by each client is ResNet18 \cite{he2016deep}. The client's local training process uses stochastic gradient descent with a learning rate of $10^{-4}$. For attack optimization, we use the Adam optimizer \cite{kingma2014adam} with a learning rate of $0.1$, together with the cosine distance loss defined in \eqref{eq_loss_cosine}, total variation regularization, and batch normalization regularization. Following the label inference methods in \cite{geng2023improved,yin2021see}, we assume that the labels are known. \Cref{tab_gradinv_scenario} lists the simulation settings for the data reconstruction attacks.

\begin{table}[!ht]
	\centering
	\caption{Simulation settings for the attacks.}
	\label{tab_gradinv_scenario}
	\begin{tabular}{cc}
		\hline
		\textbf{Client's dataset} & ImageNet \\
		\textbf{Client's neural network} & ResNet18 \\
		\textbf{Attack optimizer} & Adam \\
		\textbf{Attack learning rate} & 0.1 \\
		\textbf{Attack iterations} & 10,000 \\
		\hline
	\end{tabular}
\end{table}

As discussed in \Cref{sec_different_EB}, challenging FedAvg scenarios with $E > 1$ and $B > 1$ require an approximation strategy. For iDLG, we adapt the original approach by using random mini-batch orders across epochs. Our AWA method leverages \eqref{eq_model_update_approximate} to obtain approximate intermediate model updates and reconstruct the client's data using the first approximate epoch-wise update. GENG adopts a similar interpolation idea. However, it does not address the scenario with multiple mini-batches and lacks a layer-wise weighting mechanism. AGIC approximates the received model update by performing a single step of full-batch gradient descent on a combined batch that aggregates all mini-batches used in the client's local training (of size $N \times E$), which may lead to excessively high-dimensional dummy inputs during optimization. In all experiments, iDLG and GENG employ an unweighted loss function, whereas AGIC uses a weighted loss function with empirically chosen weights for convolutional and fully connected layers. In contrast, our AWA method uses an enhanced weighted loss whose layer weights are determined by Bayesian optimization.

\par\smallskip
\noindent\textbf{Evaluation metrics.}
To evaluate attack effectiveness and reconstruction quality, we employ three widely used metrics that quantify the difference between the reconstructed data and the ground-truth data: Learned Perceptual Image Patch Similarity (LPIPS) \cite{zhang2018unreasonable}, Peak Signal-to-Noise Ratio (PSNR) \cite{sara2019image}, and the Structural Similarity Index Measure (SSIM) \cite{wang2004imagea}. These metrics are standard and well-established indicators of image reconstruction quality and have been widely adopted in evaluations of data reconstruction attacks, see \cite{geiping2020inverting,wei2020framework,geng2023improved,xu2022agic}.

\subsection{Parameter Tuning via Bayesian Optimization}
In our AWA method, the parameter vector $Q$ defined in \eqref{eq_Q} is optimized via Bayesian optimization. Specifically, we employ a Gaussian process surrogate model as described in \eqref{eq_GP_posterior} and use the expected improvement criterion \eqref{eq_EI} as the acquisition function. The Bayesian optimization objective is the cosine distance between the reconstructed model update and the target model update, as defined in \eqref{eq_obj_func_BO_f_Q}.

Instead of tuning $Q$ for each attack, we perform the optimization once in a moderate-scale scenario ($N = 16$, $E = 32$, $B = 4$) and subsequently apply the optimized $Q^*$ across all scenarios. Our numerical results in \Cref{tab_compare_methods} demonstrate that this strategy consistently yields improved performance in both small- and large-scale scenarios while significantly reducing computational overhead.

\begin{figure}[!ht]
	\centering
	\includegraphics[width=0.43\textwidth]{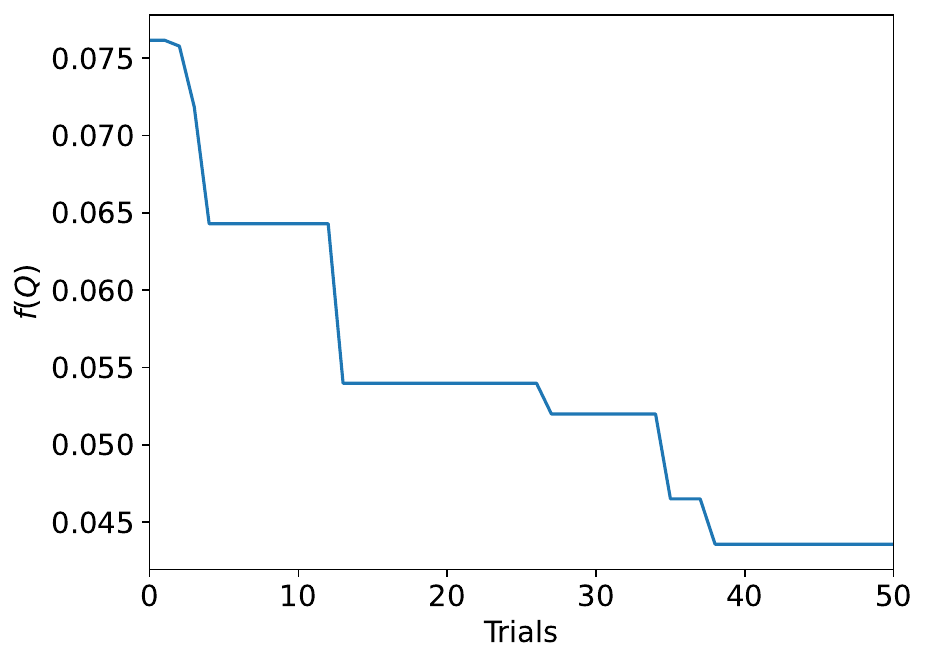}
	\caption{Cumulative minimum of Bayesian optimization.}
	\label{fig_loss_BO}
\end{figure}

Specifically, we perform Bayesian optimization for $N_{\mathrm{BO}} = 50$ iterations, with the first $n = 12$ iterations used to initialize the observation set $\mathcal{O}$. The search ranges and the optimized values for each component of $Q$ are summarized in \Cref{tab_search_ranges_BO}. The cumulative minimum loss $f(Q)$ over 50 trials is shown in \Cref{fig_loss_BO}, illustrating that Bayesian optimization consistently discovers lower $f(Q)$ values as the number of trials increases. The final tuned parameters obtained by Bayesian optimization for our AWA method are reported in the last column of \Cref{tab_search_ranges_BO}.

\begin{table}[!ht]
	\centering
	\caption{Search ranges and tuned values of $Q$ obtained by Bayesian optimization.}
	\label{tab_search_ranges_BO}
	\begin{tabular}{c c c}
		\hline
		\textbf{Parameter} & \textbf{Search range} & \textbf{Tuned value} \\
		\hline
		$q_{\mathrm{cv}}$ & [1, 1000] & 694.99 \\
		$q_{\mathrm{bn}}$ & [1, 1000] & 898.67 \\
		$q_{\mathrm{fc}}$ & [1, 1000] & 441.64 \\
		$q_{\mathrm{en}}$ & [1, 1000] & 789.51 \\
		$p_{\mathrm{mean}}$ & [0, 0.5] & 0.497 \\
		$p_{\mathrm{var}}$ & [0, 0.5] & 0.314 \\
		\hline
	\end{tabular}
\end{table}

\subsection{Results of the Image Data Reconstruction Attacks}
To verify the feasibility and effectiveness of our proposed AWA method, we evaluate it on ImageNet-based data reconstruction attacks and compare it with state-of-the-art methods, including iDLG \cite{zhao2020idlg}, AGIC \cite{xu2022agic}, and GENG \cite{geng2023improved}, under various FedAvg scenarios.

\begin{figure}[!t]
	\centering
	\begin{subfigure}[b]{0.49\textwidth}
		\includegraphics[width=\textwidth]{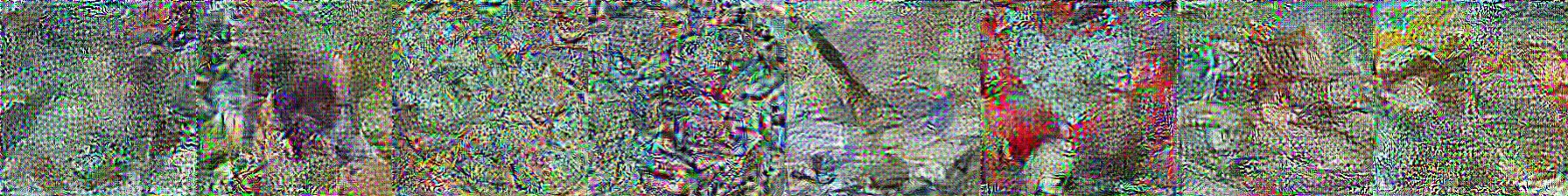}
		\vspace{-1.5em}
		\caption{iDLG \cite{zhao2020idlg}; LPIPS: 0.994, PSNR: 8.57, SSIM: 0.018.}
		\label{fig_compare_idlg}
	\end{subfigure}\\
	\vspace{3pt}
	\begin{subfigure}[b]{0.49\textwidth}
		\includegraphics[width=\textwidth]{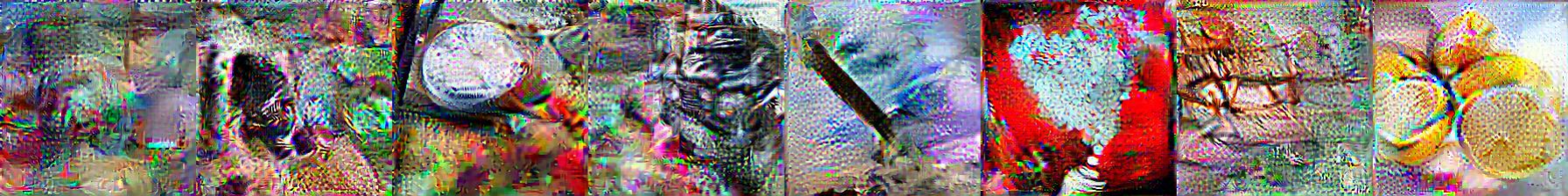}
		\vspace{-1.5em}
		\caption{AGIC \cite{xu2022agic}; LPIPS: 0.726, PSNR: 9.59, SSIM: 0.342.}
		\label{fig_compare_agic}
	\end{subfigure}\\
	\vspace{3pt}
	\begin{subfigure}[b]{0.49\textwidth}
		\includegraphics[width=\textwidth]{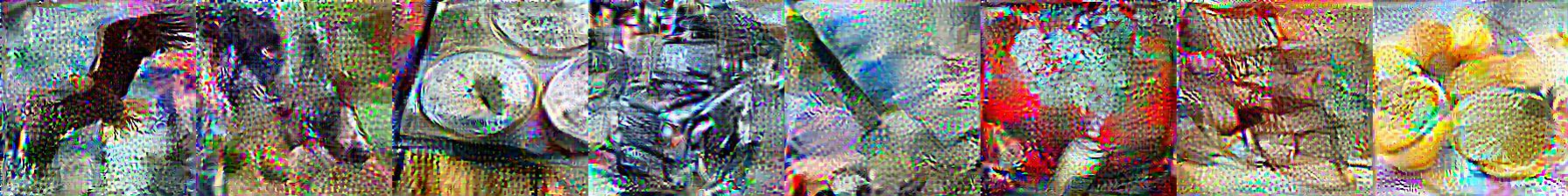}
		\vspace{-1.5em}
		\caption{GENG \cite{geng2023improved}; LPIPS: 0.672, PSNR: 10.71, SSIM: 0.468.}
		\label{fig_compare_geng}
	\end{subfigure}\\
	\vspace{3pt}
	\begin{subfigure}[b]{0.49\textwidth}
		\includegraphics[width=\textwidth]{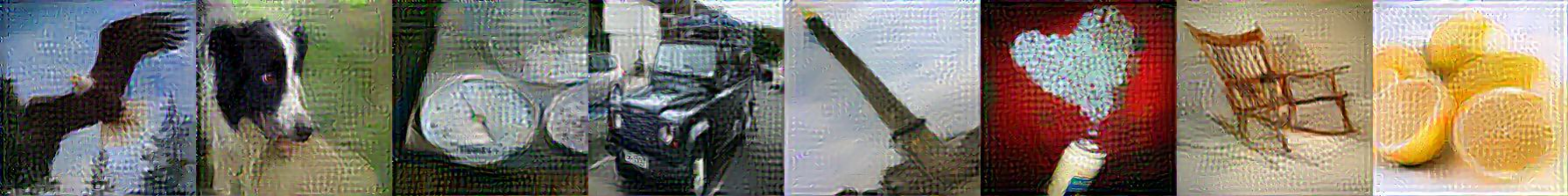}
		\vspace{-1.5em}
		\caption{AWA (ours); LPIPS: 0.508, PSNR: 15.11, SSIM: 0.800.}
		\label{fig_compare_awa}
	\end{subfigure}\\
	\vspace{3pt}
	\begin{subfigure}[b]{0.49\textwidth}
		\includegraphics[width=\textwidth]{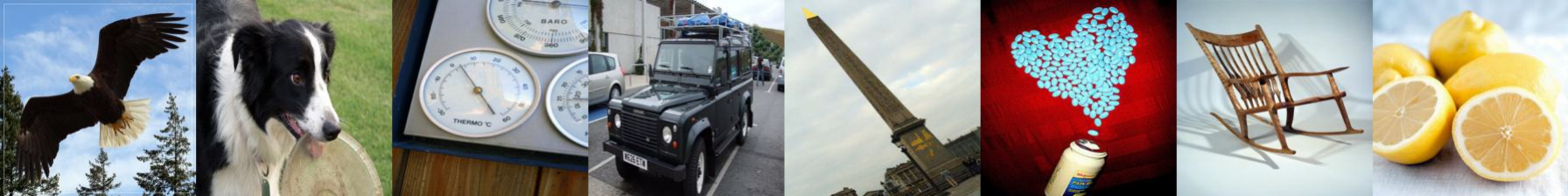}
		\vspace{-1.5em}
		\caption{Ground truth.}
		\label{fig_compare_gt1}
	\end{subfigure}
	\caption{Comparison results for attacking a FedAvg scenario ($N = 8$, $E = 32$, $B = 1$). More scenarios and quantitative metrics are summarized in Table~\ref{tab_compare_methods}.}
	\label{fig_compare_full}
\end{figure}

\begin{table}[!h]
	\centering
	\caption{Comparison of four data reconstruction attack methods across various FedAvg scenarios ($N$ images, $E$ epochs, $B$ mini-batches per epoch).}
	\label{tab_compare_methods}
	\begin{tabular}{c|c|cccc}
		\hline
		\textbf{Cases} & \textbf{Metrics} & \textbf{iDLG} & \textbf{AGIC} & \textbf{GENG} & \textbf{AWA (ours)} \\ \hline
		\multirow{3}{*}{\textbf{\begin{tabular}{c}$N = 8$\\$E = 8$\\$B = 4$\end{tabular}}} & PSNR($\uparrow$) & 8.57 & 11.13 & 10.31 & \textbf{13.49} \\
		& SSIM($\uparrow$) & 0.019 & 0.510 & 0.414 & \textbf{0.642} \\
		& LPIPS~($\downarrow$) & 0.992 & 0.659 & 0.690 & \textbf{0.573} \\
		\hline
		\multirow{3}{*}{\textbf{\begin{tabular}{c}$N = 8$\\$E = 16$\\$B = 4$\end{tabular}}} & PSNR($\uparrow$) & 8.57 & 10.47 & 10.07 & \textbf{12.42} \\
		& SSIM($\uparrow$) & 0.018 & 0.362 & 0.345 & \textbf{0.561} \\
		& LPIPS~($\downarrow$) & 0.993 & 0.703 & 0.723 & \textbf{0.611} \\
		\hline
		\multirow{3}{*}{\textbf{\begin{tabular}{c}$N = 16$\\$E = 8$\\$B = 4$\end{tabular}}} & PSNR($\uparrow$) & 9.20 & 10.82 & 11.08 & \textbf{12.13} \\
		& SSIM($\uparrow$) & 0.038 & 0.362 & 0.368 & \textbf{0.569} \\
		& LPIPS~($\downarrow$) & 0.992 & 0.687 & 0.700 & \textbf{0.646} \\
		\hline
		\multirow{3}{*}{\textbf{\begin{tabular}{c}$N = 16$\\$E = 16$\\$B = 4$\end{tabular}}} & PSNR($\uparrow$) & 9.21 & 10.45 & 10.31 & \textbf{12.33} \\
		& SSIM($\uparrow$) & 0.039 & 0.318 & 0.311 & \textbf{0.597} \\
		& LPIPS~($\downarrow$) & 0.997 & 0.707 & 0.702 & \textbf{0.642} \\
		\hline
		\multirow{3}{*}{\textbf{\begin{tabular}{c}$N = 32$\\$E = 8$\\$B = 4$\end{tabular}}} & PSNR($\uparrow$) & 9.26 & 10.16 & 9.65 & \textbf{12.21} \\
		& SSIM($\uparrow$) & 0.030 & 0.311 & 0.237 & \textbf{0.578} \\
		& LPIPS~($\downarrow$) & 1.084 & 0.717 & 0.732 & \textbf{0.636} \\
		\hline
		\multirow{3}{*}{\textbf{\begin{tabular}{c}$N = 32$\\$E = 16$\\$B = 4$\end{tabular}}} & PSNR($\uparrow$) & 9.27 & 9.93 & 9.53 & \textbf{12.04} \\
		& SSIM($\uparrow$) & 0.030 & 0.243 & 0.220 & \textbf{0.548} \\
		& LPIPS~($\downarrow$) & 1.046 & 0.747 & 0.744 & \textbf{0.647} \\
		\hline
		\multirow{3}{*}{\textbf{\begin{tabular}{c}$N = 64$\\$E = 32$\\$B = 4$\end{tabular}}} & PSNR($\uparrow$) & -- & -- & 9.14 & \textbf{11.87} \\
		& SSIM($\uparrow$) & -- & -- & 0.023 & \textbf{0.555} \\
		& LPIPS~($\downarrow$) & -- & -- & 0.804 & \textbf{0.680} \\
		\hline
		\multirow{3}{*}{\textbf{\begin{tabular}{c}$N = 64$\\$E = 32$\\$B = 8$\end{tabular}}} & PSNR($\uparrow$) & -- & -- & 9.14 & \textbf{10.95} \\
		& SSIM($\uparrow$) & -- & -- & 0.023 & \textbf{0.451} \\
		& LPIPS~($\downarrow$) & -- & -- & 0.822 & \textbf{0.694} \\
		\hline
		\multirow{3}{*}{\textbf{\begin{tabular}{c}$N = 128$\\$E = 32$\\$B = 4$\end{tabular}}} & PSNR($\uparrow$) & -- & -- & 9.06 & \textbf{11.67} \\
		& SSIM($\uparrow$) & -- & -- & 0.028 & \textbf{0.535} \\
		& LPIPS~($\downarrow$) & -- & -- & 0.804 & \textbf{0.676} \\
		\hline
		\multirow{3}{*}{\textbf{\begin{tabular}{c}$N = 128$\\$E = 32$\\$B = 8$\end{tabular}}} & PSNR($\uparrow$) & -- & -- & 9.06 & \textbf{10.57} \\
		& SSIM($\uparrow$) & -- & -- & 0.028 & \textbf{0.423} \\
		& LPIPS~($\downarrow$) & -- & -- & 0.816 & \textbf{0.702} \\
		\hline
	\end{tabular}
\end{table}

\subsubsection{Comparison of Different Attack Methods}

In this subsection, we evaluate AWA on the reconstruction of ImageNet images and compare it with other methods under various scenarios. Visual examples are presented in \Cref{fig_compare_full}, and quantitative metrics are summarized in \Cref{tab_compare_methods}.

As shown in \Cref{fig_compare_full}, AWA reconstructs the main objects with sharper contours and preserves richer texture details. The generalized iDLG approach fails to capture critical features, resulting in blurred or distorted outputs. Although AGIC and GENG reduce some artifacts relative to iDLG, they still lose fine structural information. In contrast, AWA's interpolation-based approximation combined with layer-wise weighted matching produces the most accurate and visually convincing reconstructions.

The numerical results in \Cref{tab_compare_methods} clearly support these visual findings. In each scenario, AWA achieves the highest PSNR and SSIM values and the lowest LPIPS scores, indicating superior pixel-level accuracy and perceptual quality. By comparison, iDLG requires re-running all $E$ local epochs at each attack step, leading to substantially longer runtimes. Similarly, AGIC's strategy of constructing a combined batch of size $N \times E$ imposes high memory demands as $N$ increases. Consequently, for $N \ge 64$, both iDLG and AGIC become impractical.

Overall, AWA consistently achieves superior reconstruction quality across varying numbers of images, epochs, and mini-batches, demonstrating robust and scalable performance in diverse FL scenarios.

\begin{figure*}[!t]
	\centering
	\begin{subfigure}[b]{0.98\textwidth}
		\includegraphics[width=\textwidth]{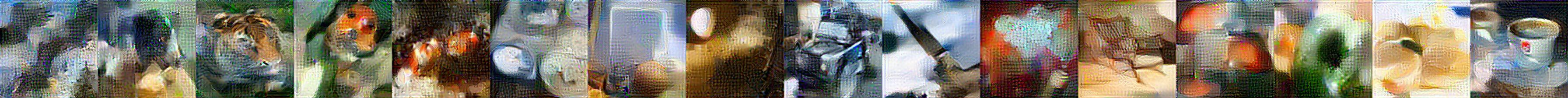}
		\vspace{-1.5em}
		\caption{AWA: approximation only; LPIPS: 0.654, PSNR: 11.23, SSIM: 0.491.}
		\label{fig_awa_approx_only}
	\end{subfigure}\\
	\vspace{3pt}
	\begin{subfigure}[b]{0.98\textwidth}
		\includegraphics[width=\textwidth]{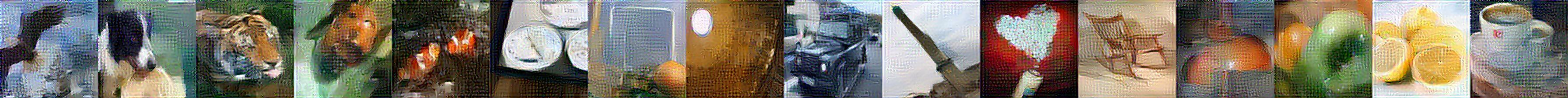}
		\vspace{-1.5em}
		\caption{AWA: approximation with linear weights; LPIPS: 0.570, PSNR: 13.87, SSIM: 0.678.}
		\label{fig_awa_linear}
	\end{subfigure}\\
	\vspace{3pt}
	\begin{subfigure}[b]{0.98\textwidth}
		\includegraphics[width=\textwidth]{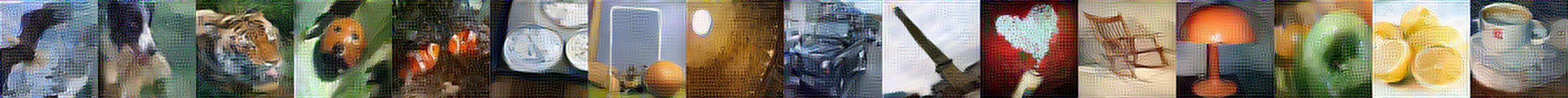}
		\vspace{-1.5em}
		\caption{AWA: approximation with linear and enhanced weights; LPIPS: 0.551, PSNR: 14.16, SSIM: 0.710.}
		\label{fig_awa_full}
	\end{subfigure}\\
	\vspace{3pt}
	\begin{subfigure}[b]{0.98\textwidth}
		\includegraphics[width=\textwidth]{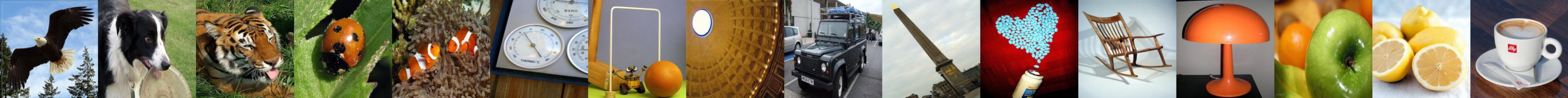}
		\vspace{-1.5em}
		\caption{Ground truth.}
		\label{fig_compare_gt2}
	\end{subfigure}\\
	\caption{Effect of layer-wise weighting in our AWA method.}
	\label{fig_awa_woQ}
\end{figure*}

\subsubsection{Effect of Layer-wise Weighting in AWA}
To assess the individual contributions of the approximation step and the weighting scheme in AWA, we compare three settings:
\begin{enumerate}
	\item \textbf{Approximation only.}
		We apply the epoch-wise interpolation in \eqref{eq_model_update_approximate} and set all layer weights to one, ignoring the linearly increasing weights and any error-driven weight enhancement.
	\item \textbf{Approximation with linear weights.}
		We keep the interpolation from the previous setting and add linearly increasing weights across layers according to \eqref{eq_q_max_3kinds}. However, we do not include the error-driven enhancement, i.e., $p_{\mathrm{mean}}=p_{\mathrm{var}}=0$.
	\item \textbf{Approximation with linear and enhanced weights.}
		This is the full AWA method. It uses both the layer-wise linearly increasing weights in \eqref{eq_q_max_3kinds} and the error-driven weight enhancement for high-error layers as in \eqref{eq_weight_aug}.
\end{enumerate}
\Cref{fig_awa_woQ} compares the reconstructions for 16 ImageNet images trained over 16 local epochs using a single mini-batch of size 16. Since all three settings use the same interpolation step, this ablation isolates how much of the improvement comes from the weighting design after approximation. The approximation-only baseline recovers only coarse shapes, leading to blurred edges, low PSNR/SSIM scores, and high LPIPS values. Adding linearly increasing weights across network layers significantly sharpens the main contours and improves all three metrics. For instance, the obelisk is reconstructed with noticeably straighter edges. Quantitatively, compared with approximation only, linear weighting improves PSNR by 23.5\% and SSIM by 38.1\%, while reducing LPIPS by 12.8\%. Relative to linear weighting alone, the enhanced weighting further improves PSNR by 2.1\% and SSIM by 4.7\%, while reducing LPIPS by 3.3\%. These results indicate that interpolation is the enabling step, most of the additional gain comes from layer-wise linear weighting, and the error-driven enhancement provides a smaller but consistent refinement in multi-epoch scenarios.

\begin{figure*}[!h]
	\centering
	\begin{subfigure}[b]{0.98\textwidth}
		\includegraphics[width=\textwidth]{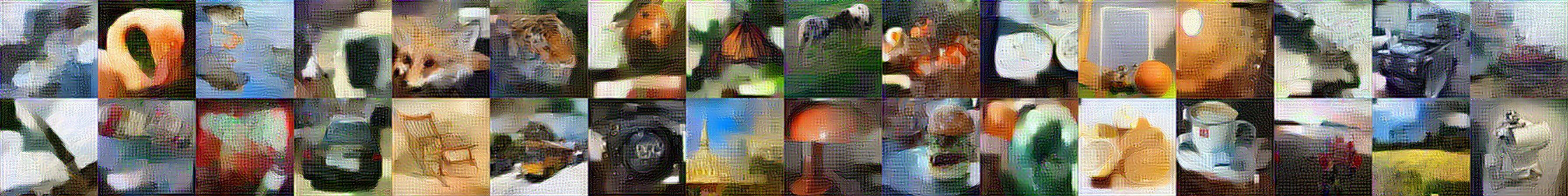}
		\vspace{-1.5em}
		\caption{$E = 32$ epochs; LPIPS: 0.659, PSNR: 11.82, SSIM: 0.528.}

	\end{subfigure}\\
	\vspace{3pt}
	\begin{subfigure}[b]{0.98\textwidth}
		\includegraphics[width=\textwidth]{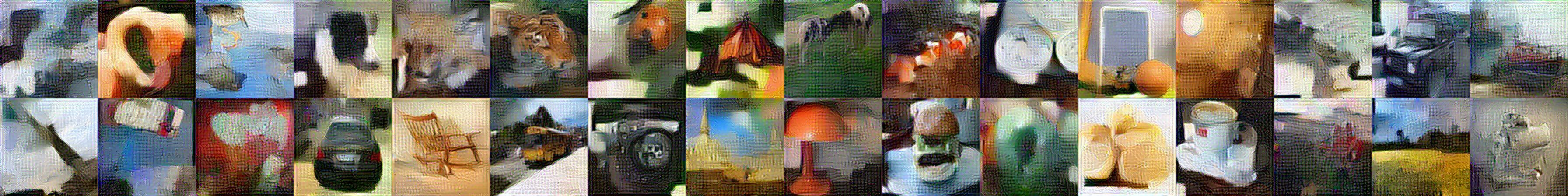}
		\vspace{-1.5em}
		\caption{$E = 16$ epochs; LPIPS: 0.647, PSNR: 12.04, SSIM: 0.548.}

	\end{subfigure}\\
	\vspace{3pt}
	\begin{subfigure}[b]{0.98\textwidth}
		\includegraphics[width=\textwidth]{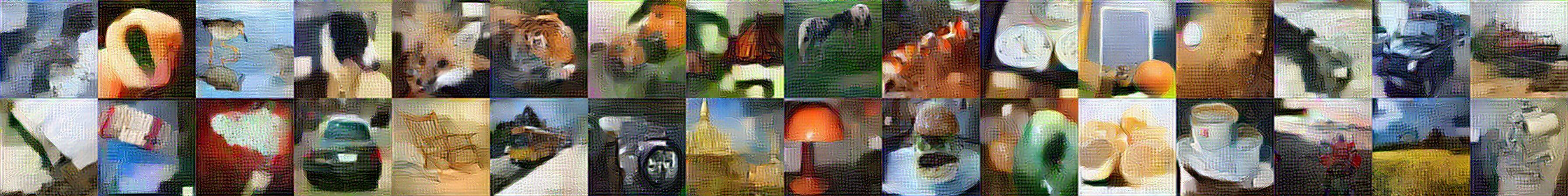}
		\vspace{-1.5em}
		\caption{$E = 8$ epochs; LPIPS: 0.636, PSNR: 12.21, SSIM: 0.578.}

	\end{subfigure}\\
	\vspace{3pt}
	\begin{subfigure}[b]{0.98\textwidth}
		\includegraphics[width=\textwidth]{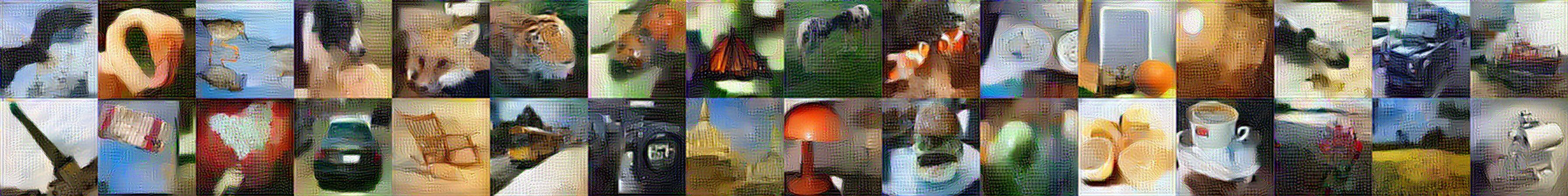}
		\vspace{-1.5em}
		\caption{$E = 4$ epochs; LPIPS: 0.633, PSNR: 12.40, SSIM: 0.600.}

	\end{subfigure}\\
	\vspace{3pt}
	\begin{subfigure}[b]{0.98\textwidth}
		\includegraphics[width=\textwidth]{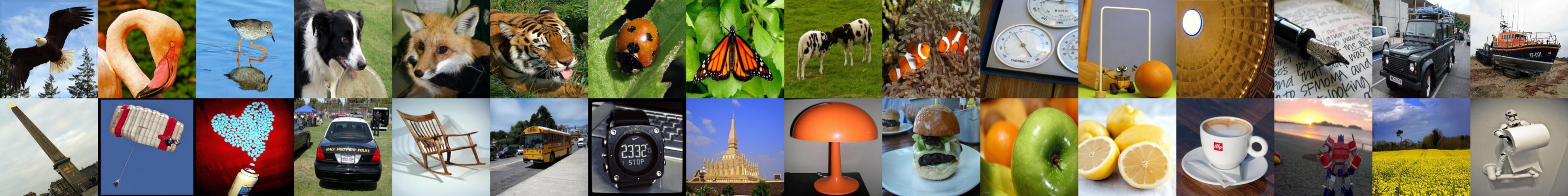}
		\vspace{-1.5em}
		\caption{Ground truth.}

	\end{subfigure}\\
	\caption{Effect of increasing the number of local epochs on the reconstruction results achieved by our AWA method.}
	\label{fig_awa_vary_E}
\end{figure*}

\begin{figure*}[!h]
	\centering
	\begin{subfigure}{0.49\textwidth}
		\centering
		\includegraphics[width=\textwidth]{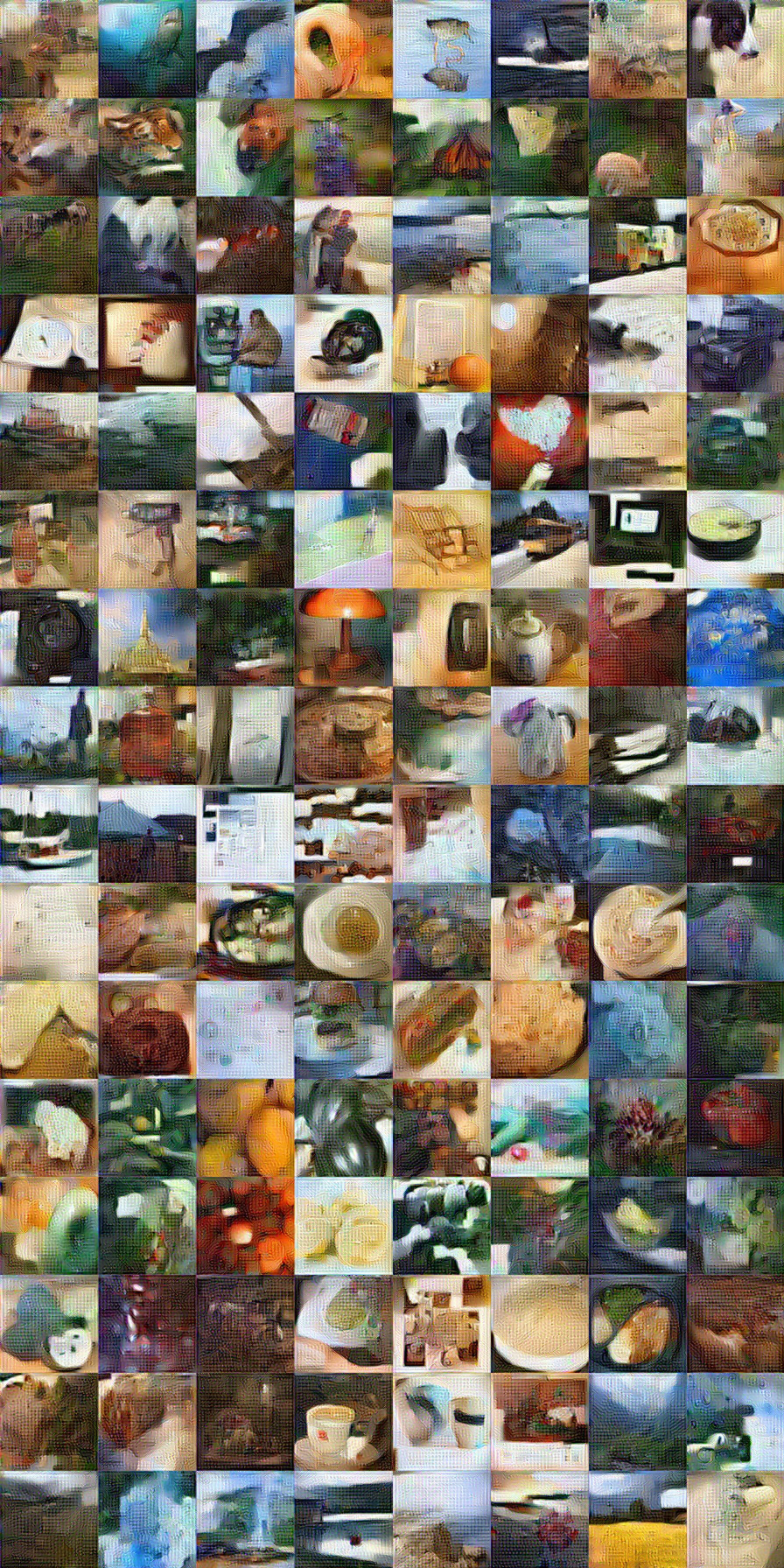}
		\caption{AWA (ours); LPIPS: 0.680, PSNR: 11.799, SSIM: 0.546.}
		\label{fig_awa_large_scale}
	\end{subfigure}
	\hfill
	\begin{subfigure}{0.49\textwidth}
		\centering
		\includegraphics[width=\textwidth]{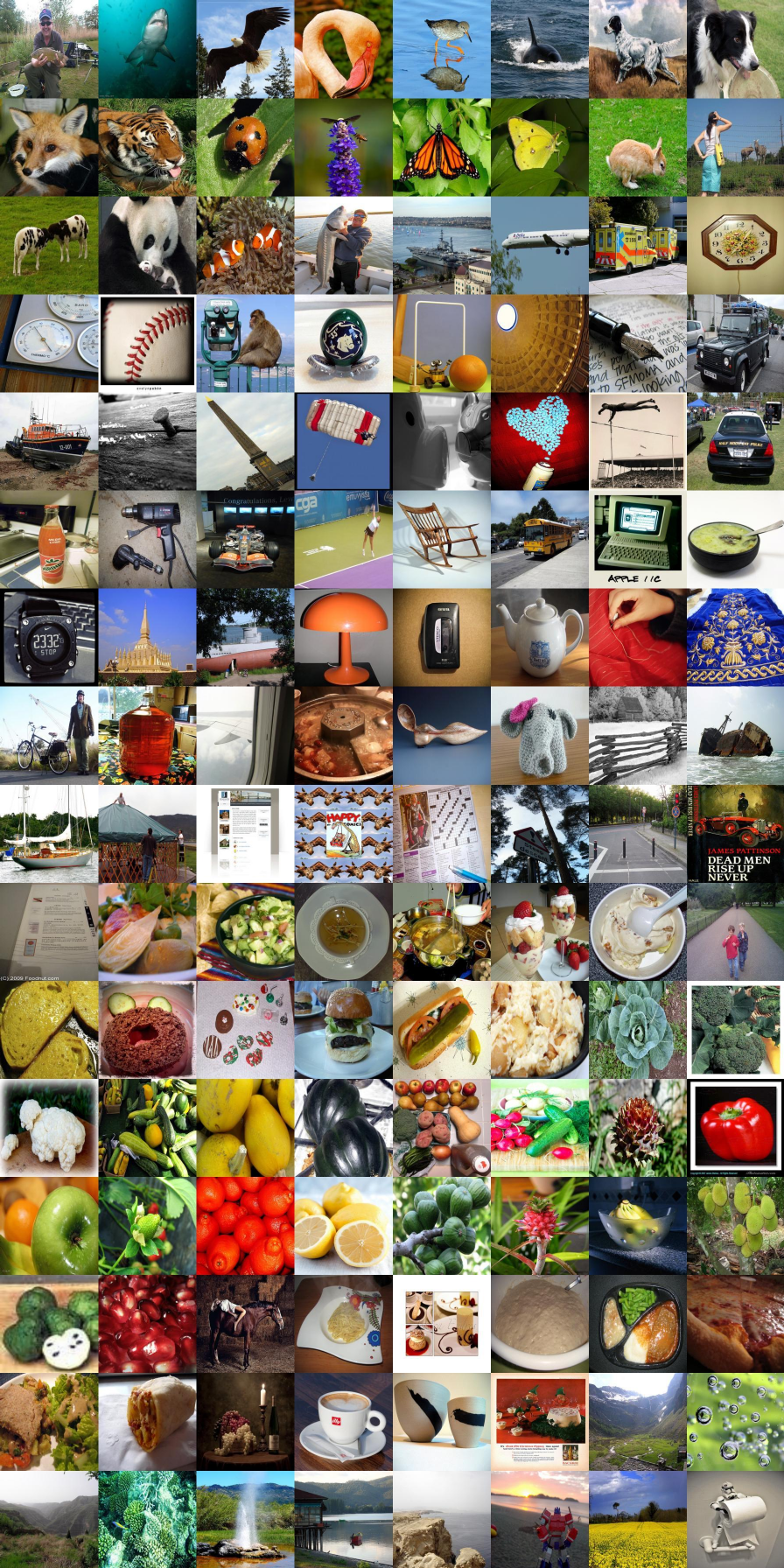}
		\caption{Ground truth.}
		\label{fig_gt_large_scale}
	\end{subfigure}
	\caption{Results of our AWA method on 128 ImageNet images trained for 64 local epochs with a mini-batch size of 32.}
	\label{fig_awa_128_64}
\end{figure*}

\subsubsection{Effect of the Number of Epochs}

We further analyze the impact of the number of local training epochs~$E$ on reconstruction performance. In a scenario with $N = 32$ images split into $B = 4$ mini-batches, we vary $E$ and present the results in \Cref{fig_awa_vary_E}. As $E$ increases, the attack becomes more challenging because of stronger model-update aggregation and repeated data shuffling. Nevertheless, AWA maintains high reconstruction quality across all tested values of $E$, with only a moderate decline in the metrics. This demonstrates the scalability and robustness of our interpolation-based approximation.

\subsubsection{Case Study of a Large-Scale Attack Scenario}
Finally, we evaluate AWA on a large-scale setup with 128 ImageNet images trained for 64 local epochs and split into four mini-batches per epoch, each with a mini-batch size of 32. To handle the added complexity, we increase the number of attack iterations from 10,000 to 50,000. As shown in \Cref{fig_awa_128_64}, AWA can still faithfully reconstruct the main visual structures and recover object categories, demonstrating its robustness in FL scenarios with large local datasets and extensive local training. Overall, these results further validate the effectiveness of AWA across diverse and realistic FedAvg scenarios.

\section{Conclusions}
\label{sec_conclusions}
The privacy benefits of FL are threatened by recently developed data reconstruction attacks. In this paper, we formulated the attack as an inverse problem, iteratively reconstructing the client's private training data by solving an optimization problem. To attack the widely used FedAvg algorithm, we proposed an interpolation-based approximation method, in which the intermediate model updates corresponding to each local epoch are effectively estimated via parameter interpolation. Furthermore, we introduced a layer-wise weighted loss function to substantially enhance reconstruction quality. By systematically assigning appropriate weights to model updates across different neural network layers via Bayesian optimization, our approach recovers training data with higher fidelity than existing state-of-the-art methods. Notably, our weighting strategy is architecture-aware, ensuring broad applicability rather than being restricted to a specific network design. Extensive numerical results demonstrated that our approximate and weighted data reconstruction attack is highly effective in realistic FedAvg scenarios, exposing critical privacy vulnerabilities in current FL systems. The ability to faithfully reconstruct raw data from leaked model updates underscores an immediate need for robust defense mechanisms. Future research will focus on extending these empirical evaluations to diverse architectures, developing rigorous countermeasures, and strengthening the security guarantees of FL frameworks to mitigate the severe risks associated with such privacy breaches.

\section*{Acknowledgments}
\vspace{-1pt}
The authors' names are listed in alphabetical order by family name to signify equal contributions.
Z. Wang was supported by the European Union's Horizon Europe MSCA project Modelling and control of flexible structures interacting with fluids (ModConFlex; grant agreement No. 101073558; call HORIZON-MSCA-2021-DN-01-01; project DOI: \href{https://doi.org/10.3030/101073558}{10.3030/101073558}).
Y. Song was supported by the NTU Start-Up Grant.
E. Zuazua was partially supported by the European Research Council (ERC) under the European Union's Horizon 2030 research and innovation programme (grant agreement NO: 101096251-CoDeFeL); the Alexander von Humboldt Professorship program; the European Union's Horizon Europe MSCA project  (ModConFlex; grant agreement No. 101073558); the Transregio 154 Project "Mathematical Modelling, Simulation and Optimization Using the Example of Gas Networks" of the DFG; the AFOSR 24IOE027 project; the SURE-AI Norwegian Centre for Sustainable, Risk-Averse, and Ethical AI grant 357482, Research Council of Norway;  by the Grant PID2023-146872OB-I00-DyCMaMod of MICIU (Spain) and by the COST Actions CA24122 - Multiscale Stochastics, Patterns, and Analysis of Combinatorial Environments and  CA24136 - Interactions between Control Theory and Machine Learning. 

\printbibliography
\vfill
\end{document}